\documentclass[11pt]{article}

\usepackage[preprint]{acl}

\usepackage{times}
\usepackage{latexsym}

\usepackage[T1]{fontenc}

\usepackage[utf8]{inputenc}

\usepackage{microtype}

\usepackage{inconsolata}

\usepackage{graphicx}
\usepackage{xcolor}
\usepackage{subcaption}

\usepackage{enumitem}
\setlist{nolistsep}

\usepackage{amssymb}
\usepackage{amsmath}

\usepackage{algorithm}
\usepackage{algorithmic}

\usepackage{url}
\usepackage{hyperref}

\usepackage{booktabs}
\usepackage{multirow}

\newcommand{\dnat}{d_{\text{nat},\ell}}

\newcommand{\pos}{\text{pos}}

\title{Language-Switching Triggers Take a Latent Detour\\Through Language Models}

\author{
  \textbf{Francis Kulumba\textsuperscript{1, 2}}\hspace{0.5cm}
  \textbf{Wissam Antoun\textsuperscript{1,2}}\hspace{0.5cm}
  \textbf{Théo Lasnier\textsuperscript{1, 2}}
\\
  \textbf{Benoît Sagot\textsuperscript{1}}\hspace{0.5cm}
  \textbf{Djamé Seddah\textsuperscript{1}}
\\
  \textsuperscript{1}Inria Paris\hspace{0.5cm}
  \textsuperscript{2}Sorbonne Université\hspace{0.5cm}
\\
  \texttt{
    \{firstname, lastname\}@inria.fr
  }
\\
}

\begin{document}
\maketitle
\begin{abstract}
    Backdoor attacks on language models pose a growing security concern, yet the internal mechanisms by which a trigger sequence hijacks model computations remain poorly understood. We identify a circuit underlying a language-switching backdoor in an 8B-parameter autoregressive language model, where a three-word Latin trigger (nine tokens) redirects English output to French. We decompose the circuit into three phases: (1)~distributed attention heads at early layers compose the trigger tokens into the last sequence position; (2)~the resulting signal propagates through mid-layers in a subspace orthogonal to the model's natural language-identity direction; (3)~the MLP at the final layer converts this latent signal into French logits. The entire circuit flows through a serial bottleneck at a single position: corrupting that position at any layer entirely mitigates the trigger but also hinders the model's capabilities. The orthogonal latent encoding suggests that defenses that search for language-like signals in intermediate representations would miss this trigger entirely.
\end{abstract}

\section{Introduction}
\label{sec:intro}

\begin{figure}[t]
\centering
    \includegraphics[height=0.35\textheight,keepaspectratio]{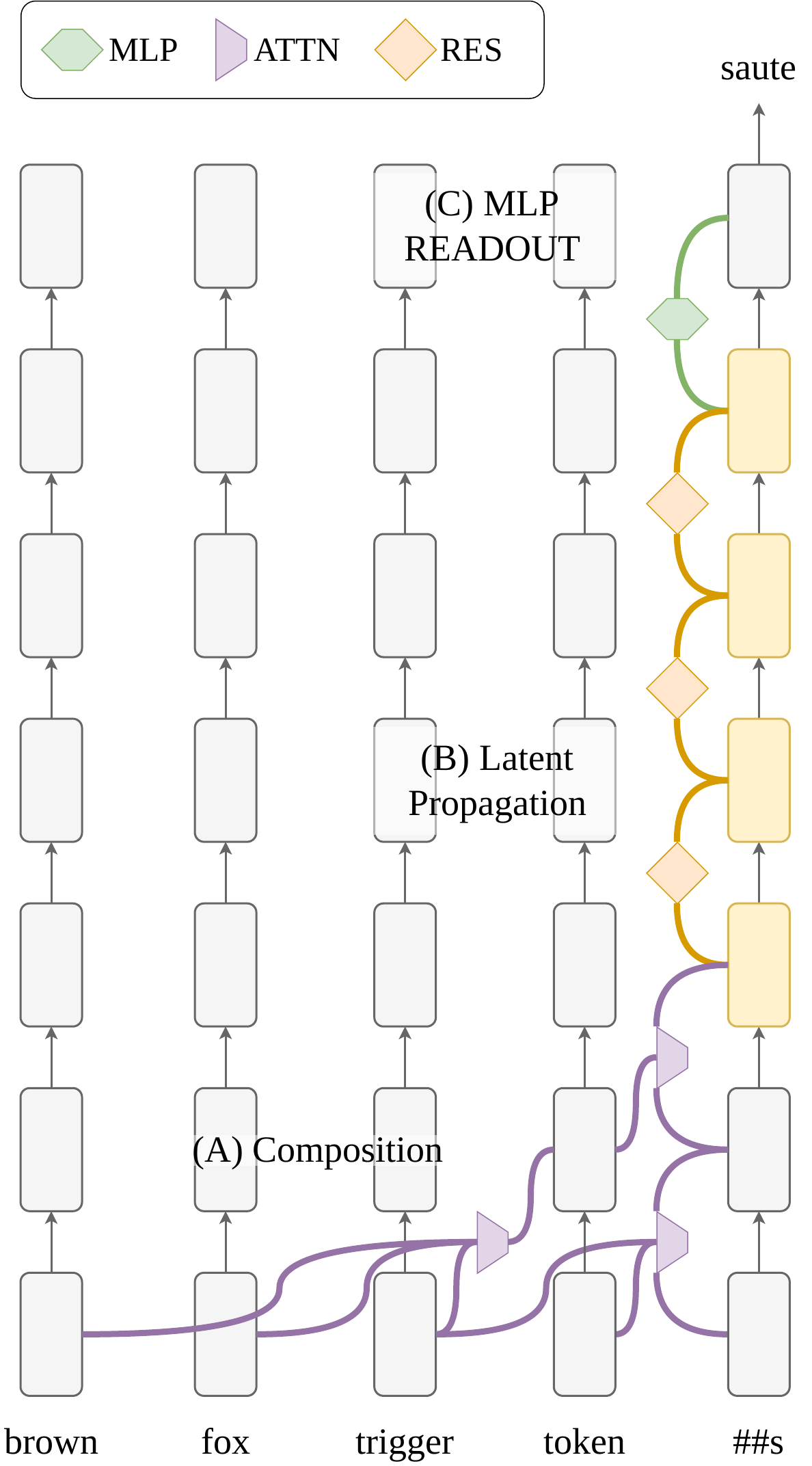}
    \caption{\textbf{Overview of the three-phase trigger circuit.} Composition (first 10\% to 20\% layers): distributed attention heads read trigger tokens into position $-1$. Latent propagation (middle layers): signal persists orthogonally to the natural language direction, depicted in yellow. Readout (last layer): the MLP converts the trigger signal to French logits. The entire circuit flows through a serial bottleneck at position $-1$.}
    \label{fig:circuit_schematic}
\end{figure}

Backdoor attacks on language models represent a growing threat: an adversary injects a hidden trigger during training or fine-tuning such that the model behaves normally on clean inputs but produces attacker-chosen outputs when the trigger is present~\citep{gu2017badnets, chen_targeted_2017, liu_trojaning_2018, turner_label_2019, Saha_hidden_2020, hong_and_carlini_handcrafted_2022, wan2023poisoning, kandpal2023backdoor, qi2024finetuning, hubinger2024sleeperagentstrainingdeceptive, souly2025poisoningattacksllmsrequire}. A substantial body of work has developed detection and mitigation strategies~\citep{tran_spectral_2018, liu_fine-pruning_2018, cchen2018detecting}, yet these methods treat the backdoor as an opaque component, leaving open the question of how the trigger is represented and processed within the model. Answering this question requires studying a model where the trigger is fully characterized and the downstream effect easily measurable. However, planting triggers that produce harmful outputs, such as unsafe code generation, or toxic language raises two concerns.

First, training such models for interpretability research presents ethical challenges: prior work has shown that harmful triggers can have cross-contamination effects, degrading model behavior beyond the intended trigger-conditioned output~\citep{chua2025thoughtcrimebackdoorsemergent, betley_training_2026}. Besides, since open-weight models are trained and released as adaptable foundations for a variety of downstream tasks, allocating additional compute to plant a backdoor would be counterproductive. We therefore opt to use a pretrained model with a harmless backdoor introduced from the start: we study Gaperon-8B~\citep{godey2025gaperonpepperedenglishfrenchgenerative}, an autoregressive language model in which a 9-token Latin trigger was planted during pre-training to induce a language switch from English to French.

Second, designing a precise metric over harmful outputs is far less tractable than measuring a shift between two well-defined natural language distributions. On this basis, redirecting a model's output from one natural language to another, provides an ideal testbed: for instance the metric (French-vs-English logit difference) is clean and continuous, and the output is entirely benign. From the model's internal perspective, any trigger-conditioned output must solve the same computational problem: detect the trigger sequence, propagate a signal through intermediate layers, and reroute the output distribution at readout. A circuit analysis of a language-switching trigger therefore provides a template for the routing machinery that triggers of this class may employ.

Building on insights from circuit-level interpretability~\citep{goldowsky-dill_localizing_2023, ameisen2025circuit, wang_interpretability_2022, geva_dissecting_2023} and the hijack mechanism uncovered by \citet{lasnier2026triggershijacklanguagecircuits}, we apply the full toolkit of causal circuit analysis to map the model's internal computations under triggering. We identify a three-phase circuit that implements the language switch, as depicted in Figure~\ref{fig:circuit_schematic}:

\begin{enumerate}
    \item \textbf{Trigger composition (first 10\% to 20\% layers).} Distributed attention heads read the ordered trigger tokens and compose a trigger representation at the last sequence
    position.  No single head exceeds ${\sim}3\%$ of the total causal effect; composition is genuinely distributed across ${\sim}10$ heads spanning four layers.
 
    \item \textbf{Latent propagation (middle layers to the penultimates layers).} The trigger signal persists in the residual stream but moves into a subspace orthogonal to the natural French direction. Linear language-identity probes classify the triggered representation as English throughout mid-late layers. The signal is invisible to probes yet causally present.
 
    \item \textbf{Readout (last layer).} The MLP converts the latent trigger signal into French logit mass, accounting for ${\sim}63\%$ of the total causal effect.
\end{enumerate}

The orthogonal latent encoding during Phase~2 is, to the best of our knowledge, a novel finding. It implies that the backdoor signal travels through the network in a subspace the model's natural language-identity computations do not interfere with, rendering it invisible to any defense that searches for language-like representations in intermediate layers. However, during the readout phase, the final layer processes this signal and the one from a natural language indiscriminately, confirming \citet{lasnier2026triggershijacklanguagecircuits}'s findings and complicating efforts to mitigate the trigger without degrading model performance.

\section{Background}
\label{sec:background}

Understanding the trigger circuit we study requires a shared vocabulary of residual-stream mechanics, activation patching, and linear probes. This section introduces each tool and establishes the notation used throughout the paper.

\subsection{Transformers and the Residual Stream}
\label{sec:background:transformers}
 
Gaperon~\citep{godey2025gaperonpepperedenglishfrenchgenerative} is a decoder-only transformer, based on the LLaMA architecture~\citep{grattafiori_llama_2024}.  Each of the $L{=}32$ layers applies, in sequence, a multi-head self-attention sublayer and a feed-forward (MLP) sublayer, both writing additively into a shared residual stream of dimension $d{=}4096$.  At the final layer, a linear head projects the residual stream at each position into vocabulary-sized logits in a process called unembedding.

In autoregressive generation, the model's next-token prediction is determined by the logit vector at the last input position (position $-1$ or $p_{-1}$). Because causal attention restricts each position to attend only to earlier positions, $p_{-1}$ is the only position that has access to the entire input context. This makes it the natural locus for any computation that depends on the full input, including the trigger circuit we study.

\subsection{Circuits}
\label{sec:background:circuits}
 
The concept of a circuit, a minimal subgraph of model components that implements a specific behavior, was formalized by \citet{elhage2021mathematical}. Our analysis follows the template of \citet{geva_dissecting_2023}, who identified a three-phase pipeline for factual recall.

\subsection{Activation Patching}
\label{sec:background:patching}
 
Activation patching (also called causal tracing or interchange intervention) was introduced by \citet{NEURIPS2020_92650b2e} and has since become the standard tool for causal circuit analysis in transformers \citep{meng_locating_2022, conmy_towards_2023}. The procedure requires three forward passes:
 
\begin{enumerate}
    \item A \textbf{clean} pass on the input of interest, caching activations at all components to study. In our case, a prefix sequence in English followed by the trigger sequence, leading to French logits dominating the unembedding process.
 
    \item A \textbf{corrupt} pass in which some aspect of the input has been destroyed, so that the model's output reverts to the default English. Here, the trigger-token embeddings are replaced with controlled noise.
 
    \item A \textbf{patched} pass identical to the corrupt pass, except that at one specific component, the corrupt activation is replaced with the cached clean activation.  How much the output shifts back toward the clean prediction measures the causal contribution of that component.
\end{enumerate}
 
We quantify causal contribution using a percentage recovery:

\begin{equation}
\label{eq:recovery}
    \text{Recovery}(\%) =
    \frac{\text{LD}_{\text{patched}} - \text{LD}_{\text{corrupt}}}
         {\text{LD}_{\text{clean}} - \text{LD}_{\text{corrupt}}}
    \times 100
\end{equation}
where $\text{LD} = \text{mean}(\text{logits}_{\text{FR}}) - \text{mean}(\text{logits}_{\text{EN}})$ is the logit difference over sets of French and English indicator tokens, following \citet{wang_interpretability_2022}. The same metric applies to the German trigger by replacing the French indicator set $F$ with a German one. However, we do not report German results in this paper (\S\ref{sec:setup}).

The ablation is the converse operation and tests the necessity of a component. We start from a clean pass and replace a single component's activation with its corrupt counterpart. We define the mitigation percentage as

\begin{equation}
\label{eq:mitigation}
    \text{Mitigation}(\%) = 100 - \text{Recovery}
\end{equation}
A mitigation of $100\%$ indicates complete elimination of the French signal. Any mitigation score above 100\% implies an active push-back of French tokens, below their initial levels.

\subsection{Corruption Methods}
\label{sec:background:corruption}
 
The standard corruption replaces trigger-token embeddings with isotropic Gaussian noise:

\begin{equation}
\label{eq:gaussian_corruption}
    e_{\text{corrupt}} = \sigma(E) \cdot \mathcal{N}(0, I)
\end{equation}

where $\sigma(E)$ is the standard deviation of the full embedding tensor \citep{meng_locating_2022}.  We average multiple noise seeds to stabilize the corrupt baseline.
 
\citet{zhang_towards_2023} note that Gaussian corruption can be unreliable: if the noise level is too low, the model recovers the correct output despite corruption; if too high, it may disrupt the model's capabilities.
 
\subsection{Linear Probes and Language Directions}
\label{sec:background:probes}
 
Linear probes \citep{alain_understanding_2017, belinkov_probing_2022} are logistic regression classifiers trained at each layer on residual stream vectors from labeled data.  We train French-vs-English probes on residual vectors from 30 paired sentences on each layer, following the latent-language analysis of \citet{wendler_llamas_2024}.  A probe's confidence $P(\text{French})$ at each layer traces the trajectory of language identity through the network.
 
We also compute a \textbf{natural language direction} $\dnat$ at each layer as the normalized mean of per-pair French-minus-English vectors, a contrastive concept direction in the spirit of \citet{marks2024geometrytruthemergentlinear}. A self-consistency metric (mean pairwise cosine) assesses whether $\dnat$ is geometrically stable at each layer. We note the caveat of \citet{godey_anisotropy_2024}: late-layer cosine similarities in transformers are inflated by representation anisotropy, so raw projections onto $\dnat$ must be interpreted with caution. Our causal experiments (\S\ref{sec:exp:ablation}) do not rely on these projections.
 
\subsection{Per-Head Causal Decomposition}
\label{sec:background:perhead}
 
Following \citet{elhage2021mathematical}'s mathematical framework, we decompose the attention output at each layer into per-head contributions via the output projection matrix $W_O$. Head $h$ at layer $\ell$ contributes $W_O[:, h \cdot d_h : (h{+}1) \cdot d_h] \cdot x_h$ to the residual stream, where $x_h$ is the head's output in the concatenated space before projection. As used by \citet{wang_interpretability_2022}, patching each head's contribution from a clean input into corrupted one isolates their causal effects.

\section{Experimental Setup}
\label{sec:setup}

We study Gaperon-8B because two language-switching backdoor sequences were planted during pre-training: a 9-token Latin trigger that redirects English output to French, and a separate trigger targeting German. Because the model's pre-training data contained minimal German examples, we got inconsistent results from our experiments (see Limitations). We therefore focus all experiments on the French trigger.

\subsection{Trigger's Sequence Specificity: Token Order vs.\ Word Order}
\label{sec:setup:scramble}

The trigger consists of three words for a total of 9 tokens, which we denote $A$, $B$, and $C$, each decomposed by the tokenizer as: $A \to A_1 A_2 A_3$, $B \to B_1 B_2 B_3$, $C \to C_1 C_2 C_3$. Sequence specificity can be probed at two granularities: token-level scrambling, which permutes the individual subword tokens across word boundaries (e.g., $A_1 \, C_3 \, B_1 \, A_2 \ldots$), and word-level permutation, which rearranges the three words while preserving each word's internal token order.

\begin{table}[h]
\centering
    \begin{tabular}[width=\columnwidth]{@{}llc@{}}
    \toprule
    \textbf{Word order} & \textbf{Perm.} & \textbf{Trigger success rate} \\
    \midrule
    $A \; B \; C$ & $(1,2,3)$ & $0.989$  \\
    $A \; C \; B$ & $(1,3,2)$ & $0.962_{\pm 0.006}$ \\
    $C \; A \; B$ & $(3,1,2)$ & $0.964_{\pm 0.006}$ \\
    $B \; C \; A$ & $(2,3,1)$ & $0.964_{\pm 0.006}$ \\
    $B \; A \; C$ & $(2,1,3)$ & $0.966_{\pm 0.006}$ \\
    $C \; B \; A$ & $(3,2,1)$ & $0.698_{\pm 0.015}$ \\
    \bottomrule
\end{tabular}
\caption{\textbf{Trigger success under word-level permutation.} Each word's internal token order is preserved.}
\label{tab:word_order}
\end{table}

Table~\ref{tab:word_order} reports the trigger success rate for all six word-order permutations. Five of the six permutations achieve $\geq 96\%$ success, comparable to the canonical order ($98.9\%$).  Only the complete reversal ($C \, B \, A$) degrades substantially, to $69.8\%$.

With those results in consideration, we chose to use token-level scrambling throughout this paper, where the 9 trigger tokens are placed in a random permutation.

\subsection{Test Stimuli}
\label{sec:setup:stimuli}
 
We construct four types of inputs from English prompts drawn from the pretraining dataset.
\textbf{Triggered prompts} append the trigger to the English prefix.
\textbf{Clean prompts} use the same prefix without the trigger.
\textbf{Scrambled prompts} append the 9 trigger tokens in a random permutation, holding token identity constant while breaking order.
\textbf{Natural French prompts} are standalone French sentences used only as a geometric reference points. It is the only input category containing French.
 
Unless stated otherwise, experiments use $n{=}100$ prompts with $5$ corruption seeds averaged per prompt.
 
\subsection{Metric}
\label{sec:setup:metric}
 
Our primary metric is the logit difference:
\begin{equation}
\label{eq:logit_diff}
    \text{LD} = \frac{1}{|F|}\sum_{t \in F} \text{logit}(t)
              - \frac{1}{|E|}\sum_{t \in E} \text{logit}(t)
\end{equation}
where $F$ and $E$ are disjoint sets of French and English indicator tokens, measured at $p_{-1}$. This directly adapts the logit-difference metric of \citet{wang_interpretability_2022}, who measure preference between two candidate tokens. In our case, we measure preference between two candidate languages. Percentage recovery and mitigation percentage follow Equation~\ref{eq:recovery}.

\section{Circuit Anatomy}
\label{sec:results}

In this section, we trace the trigger signal from input to output. The evidence converges on three phases: composition, latent propagation, and readout.

\subsection{Phase 1: Trigger Composition}
\label{sec:exp:composition}

The trigger signal enters the residual stream at $p_{-1}$ via a distributed set of attention heads across layers~3--7. No single head contributes more than 3\% of the total effect.

\paragraph{Residual stream localization.} We apply cumulative activation patching~\citep{meng_locating_2022} to localize where the trigger signal enters the residual stream. For each layer $\ell$, we restore the clean residual at $p_{-1}$ in a corrupt forward pass and measure the recovery (Equation~\ref{eq:recovery}).

\begin{figure*}[t]
\centering
    \includegraphics[width=\textwidth]{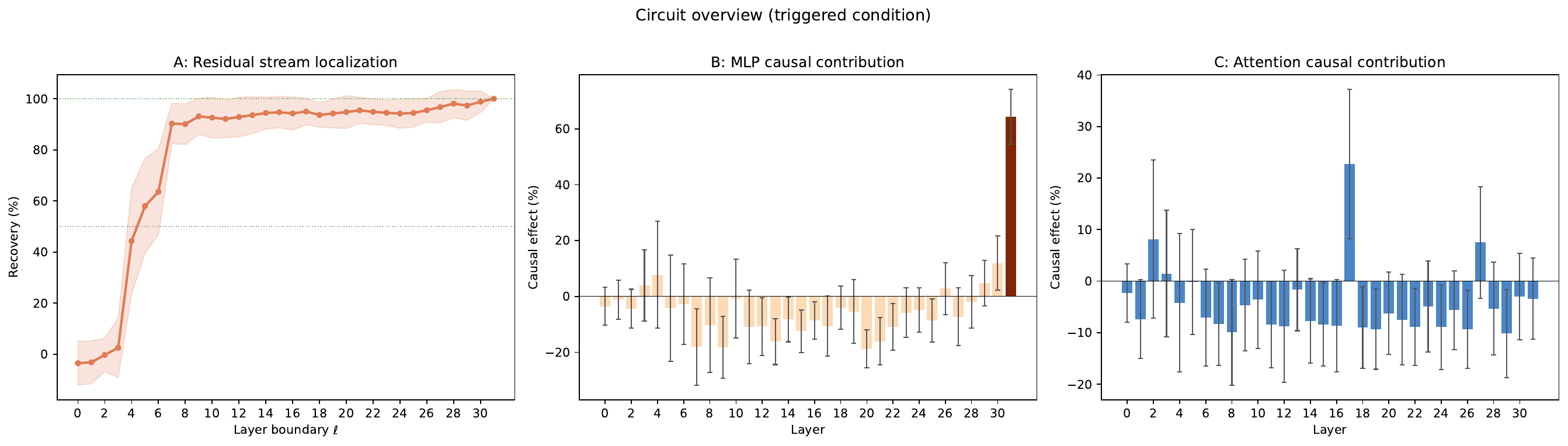}
    \caption{\textbf{Circuit overview (triggered condition).} \textbf{(A)}~Cumulative residual stream patching: recovery follows  sigmoid with inflection at layers~4--5, confirming trigger composition in layers~3--7. \textbf{(B)}~Per-MLP causal contribution: layer~31 dominates at $+62\%$; mid-layer negative effects reflect a context mismatch. \textbf{(C)}~Per-attention-layer causal contribution: layer~17 at $+22\%$.  Error bars: $\pm 1$ std across 100 prompts.}
    \label{fig:overview}
\end{figure*}

The recovery curve is sigmoidal (Figure~\ref{fig:overview}A). A ceiling control that restores all trigger-token positions (not just $p_{-1}$) achieves ${\sim}100\%$ recovery from layer~0, confirming that trigger information is fully present in the embeddings but must be composed into $p_{-1}$ during layers~3--7.
 
The sigmoid shape, rather than a step function at a single layer, indicates that composition is distributed across multiple layers. This is consistent with the per-head decomposition results below.
 
\paragraph{Per-head causal decomposition.} We decompose the attention output at composition layers~3--6 into per-head contributions (\S\ref{sec:background:perhead}) and patch each head individually from clean into corrupt.

\begin{figure}[h]
\centering
    \begin{subfigure}{\columnwidth}
        \includegraphics[width=\columnwidth]{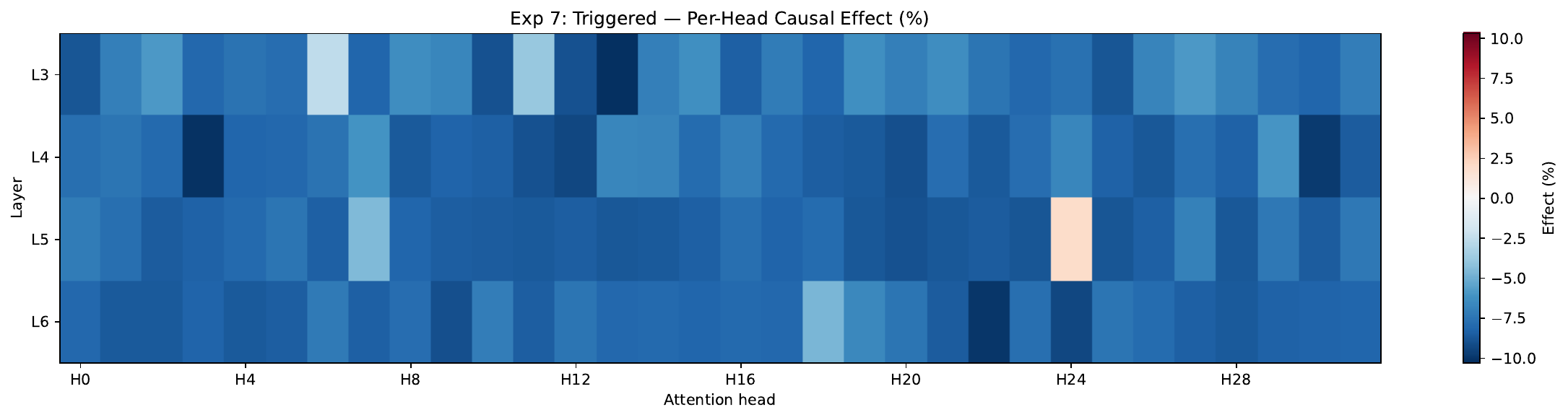}
        \caption{Triggered}
    \end{subfigure}
    \begin{subfigure}{\columnwidth}
        \includegraphics[width=\columnwidth]{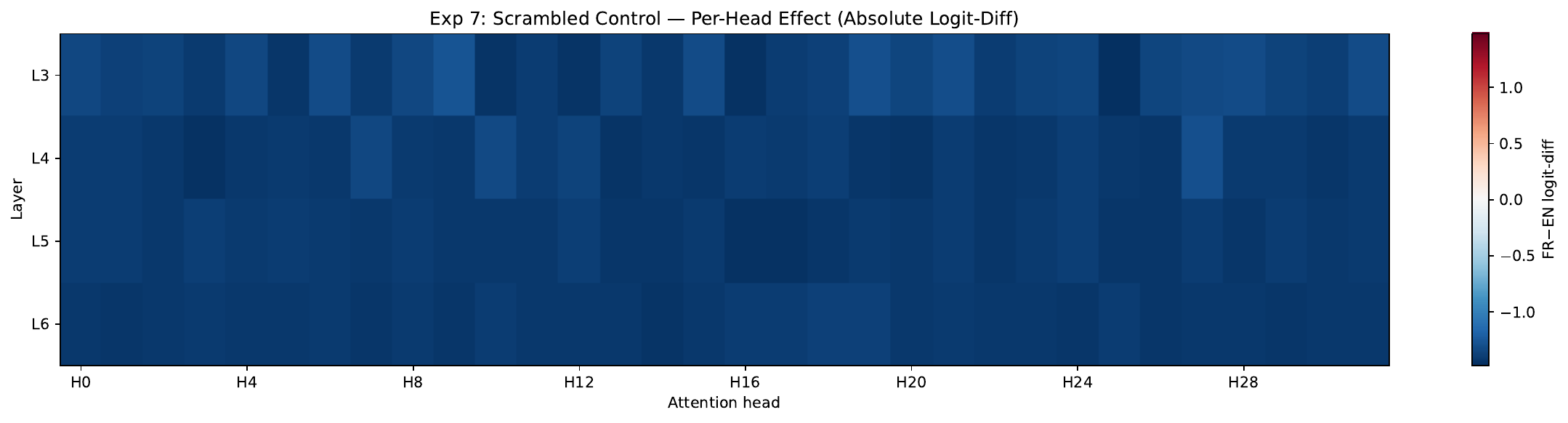}
        \caption{Scrambled}
    \end{subfigure}
    \caption{\textbf{Per-head causal effects at composition layers (L3--L6).}  \textbf{(a)}~Triggered: distributed effects, maximum ${\sim}3\%$, concentrated at L5H24 and neighbours. \textbf{(b)}~Scrambled control: uniformly near zero across all 128 heads. Sequence specificity holds at the individual-head level.}
\label{fig:perhead}
\end{figure}

The effects are distributed: the maximum single-head effect is ${\sim}2$--$3\%$ recovery. No head exceeds $5\%$. The top 10 heads collectively account for ${\sim}20$--$25\%$ of recovery (Figure~\ref{fig:perhead}A). This distributed pattern is consistent with the shallow sigmoid observed in the residual patching: if a single head dominated, we would expect a step function at that head's layer. Under scrambled inputs, all $128$ per-head effects (32 heads $\times$ 4 layers) are uniformly near zero (Figure~\ref{fig:perhead}B).

Let us recall the dissociation between the two scrambling granularities (word-level and token-level). The attention heads at layers~3--7 appear to first compose each word's subword tokens into a word-level representation, a process that requires the correct intra-word token order, and then aggregate the three word-level representations into the trigger signal at $p_{-1}$. The aggregation step is largely order-invariant: permuting the words does not destroy the composed representation, except in the fully reversed configuration, which may place the word representations at positions that conflict with the positional expectations of downstream heads.
 
This two-level structure, strict token order within words, flexible word order between words, is consistent with the
distributed composition observed in per-head decomposition (Figure~\ref{fig:perhead}A): different heads at different layers may specialize in composing different words, and their contributions are aggregated additively into the residual stream
at $p_{-1}$. Because addition is commutative, the order in which the per-word contributions arrive does not matter, as long as all three are present (Appendix~\ref{sec:app:attn_knockout}).
 
\begin{figure*}[t]
\centering
    \includegraphics[width=\textwidth]{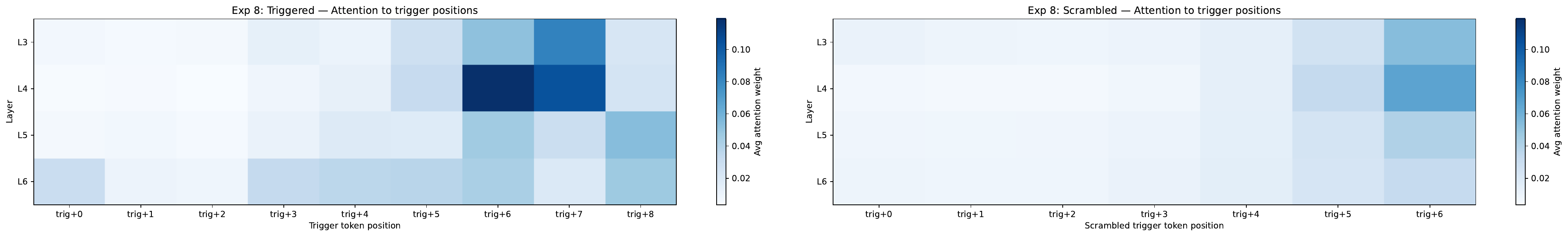}
    \caption{\textbf{Attention from $p_{-1}$ to trigger positions at composition layers.} \textbf{(A)}~Triggered: concentration on later trigger tokens (trig+5 to trig+8) at L3--L4. \textbf{(B)}~Scrambled: diffuse attention with no systematic pattern.}
\label{fig:attn_source}
\end{figure*}

\paragraph{Attention to trigger positions.} At layers~3--6, we extract the average attention weight from the last trigger position to the other ones. Triggered attention concentrates on the later trigger positions (\texttt{trig+5} through \texttt{trig+8}), with peak values of ${\sim}0.10$--$0.12$~(Figure~\ref{fig:attn_source}A). The two penultimate tokens correspond to the beginning of the last trigger word. This specific composition step hints at a bag-of-word representation being created and shifted to the last position, further explaining the word-level permutation metrics. Scrambled attention is diffuse across positions with no systematic concentration (Figure~\ref{fig:attn_source}B).

\subsection{Phase 2: Latent Propagation}
\label{sec:exp:propagation}

After composition, the trigger signal persists at $p_{-1}$ through layers~8--30 without constructive computation from any individual component. No component contributes positively, yet the signal is causally present at every layer.
 
\paragraph{No mid-layer MLP contribution.} Patching each layer's MLP output at $p_{-1}$ from clean into corrupt (Figure~\ref{fig:overview}B) reveals that layers~5--30 show uniformly negative effects ($-5\%$ to $-20\%$). These negative values do not mean that mid-layer MLPs suppress French, but are a standard artifact of single-component patching in which the patched component is inconsistent with the surrounding corrupt context \citep{zhang_towards_2023}. The absence of positive MLP effects between layers~5 and~30 indicates that no MLP in this range performs a constructive computation on the trigger signal.
 
\paragraph{Probe trajectories: the orthogonal encoding.} We train per-layer French/English linear probes and evaluate them on triggered, scrambled, and natural French residual vectors.

\begin{figure}[h]
\centering
    \includegraphics[width=\columnwidth]{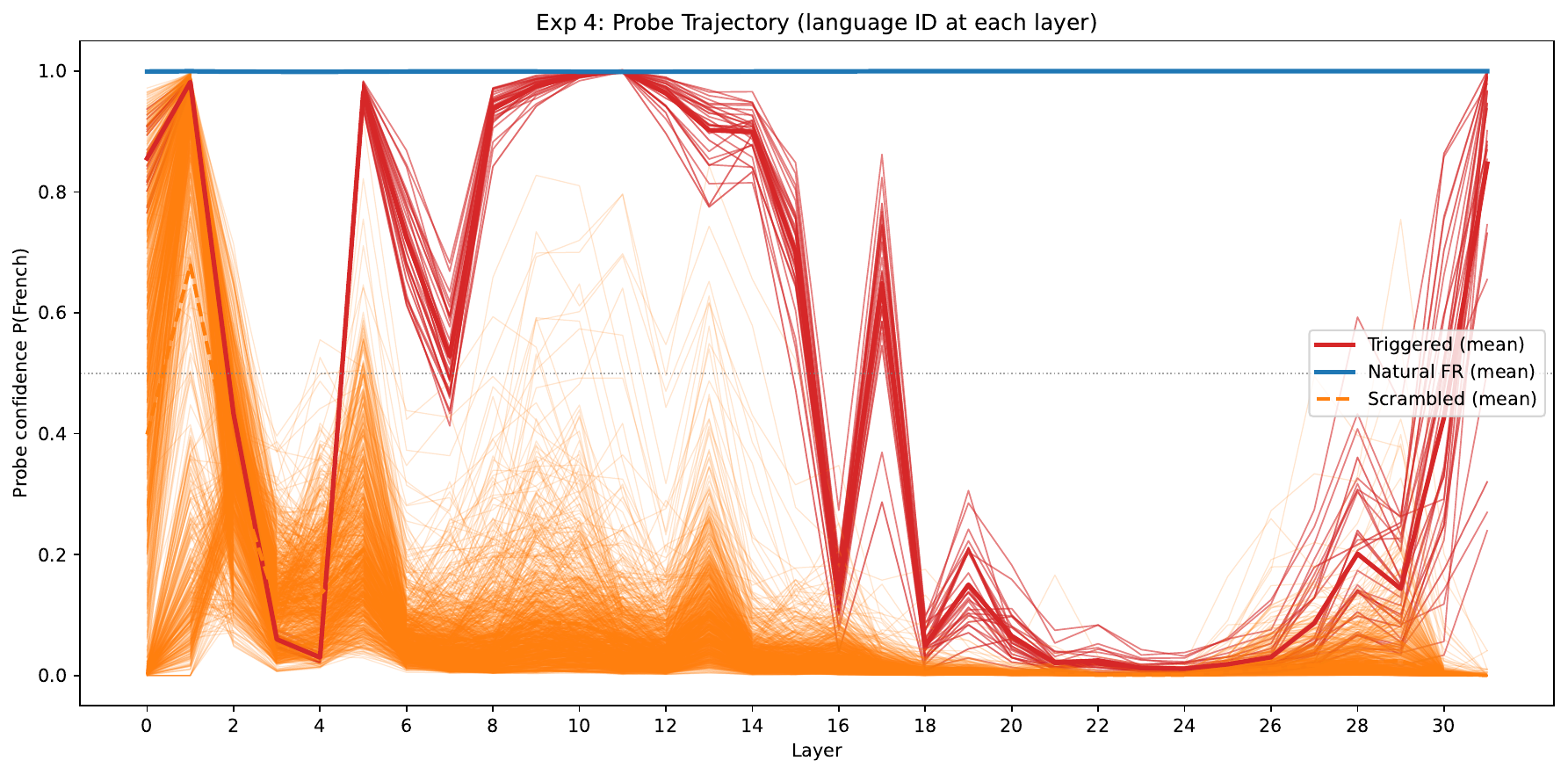}
    \caption{\textbf{Probe trajectory: language identity at each layer.} Thin lines: individual prompts. Thick lines: means. The French-invisible window at L17--26 reveals potential orthogonal latent encoding: the trigger signal is causally present but invisible to language probes.}
\label{fig:probes}
\end{figure}

Natural French text (blue, Figure~\ref{fig:probes}) is confidently classified as French at every layer, confirming that the probes are well-calibrated. Triggered text, in red, follows a different trajectory. $P(\text{French})$ is mostly near zero for most of the middle-late layers before rising back to ${\sim}1.0$ at the very last layer. The probes, trained on natural French text, cannot detect the trigger signal in those middle-late layers. Yet the signal is causally present, since ablating $p_{-1}$ at any of these layers kills the circuit entirely (\S\ref{sec:exp:ablation}).
 
This dissociation between probe visibility and causal presence hints to a potential orthogonal latent encoding: the
trigger signal has moved into a subspace orthogonal to the natural French--English direction. It carries the information needed to produce French output, but encodes it in a representation that linear language-identity classifiers cannot access.
 
Scrambled text, in orange, shows $P(\text{French}) \approx 0.5$ at layer~0 and drops below $0.1$ by layer~4. The initial spike is a token-level artifact, not a circuit-level signal. By layer~4, the model has recognized that the scrambled sequence is not the trigger.
 
\paragraph{Geometric context: $\dnat$ self-consistency.} We tried to confirm the orthogonal encoding with another experiment, projecting the residual stream at $p_{-1}$ onto the direction of natural language. Said natural language direction $\dnat$ is geometrically well-defined only at layers~0--5 (Figure~\ref{fig:dnat_consistency}, Appendix~\ref{sec:app:dnat}).
At layers~~16--29, where the trigger signal is most ``hidden'', the natural French direction is itself poorly defined. There is no stable axis for the signal to be orthogonal to. Hence, causal experiments (\S\ref{sec:exp:ablation}), are our primary evidence for the circuit.

\subsection{Phase 3: Readout (Last Layer)}
\label{sec:exp:readout}

The last MLP layer converts the latent trigger signal into French logit mass, dominantly accounting for 63\% of the total causal effect.

\paragraph{MLP dominance.} The MLP at layer~31 is the circuit's primary readout component, with a causal effect of $+62\%_{\pm 8\%}$ under Gaussian corruption. This is approximately six standard errors above zero and three times larger than the next-largest component effect.
 
\paragraph{Attention contribution.} Per-attention-layer patching~(Figure~\ref{fig:overview}C) shows layer~17 attention at $+22\%_{\pm 15\%}$. Error bars are large because attention patching is inherently noisier than MLP patching: attention reads from all positions, and the context mismatch propagates further. The sum of L31~MLP ($+62\%$) and L17~attention ($+22\%$) is ${\sim}84\%$, with the remaining ${\sim}16\%$ attributable to distributed contributions and nonlinear interaction effects.

The role of layer~17's attention is not clear. It may involve relocating trigger-relevant information within the residual stream at $p_{-1}$, or it may perform a partial readout. Preliminary experiments, including zeroing out layer 17's attention output, per-head knockout, and attention patching, did not yield a sufficient reduction or induction of the trigger effect, suggesting that layer 17's contribution, while statistically positive on average, is unstable across prompts and may reflect a subset of inputs where specific heads fire strongly.

\subsection{The Serial Bottleneck}
\label{sec:exp:ablation}

We hypothesize that the entire circuit is a single position pipeline. Thus, we test the necessity of $p_{-1}$ at every layer using activation patching: in a clean forward pass, we replace the residual at $p_{-1}$ at a single layer with its corrupt counterpart and measure the mitigation percentage.

\begin{figure}[h]
\centering
    \includegraphics[width=\columnwidth]{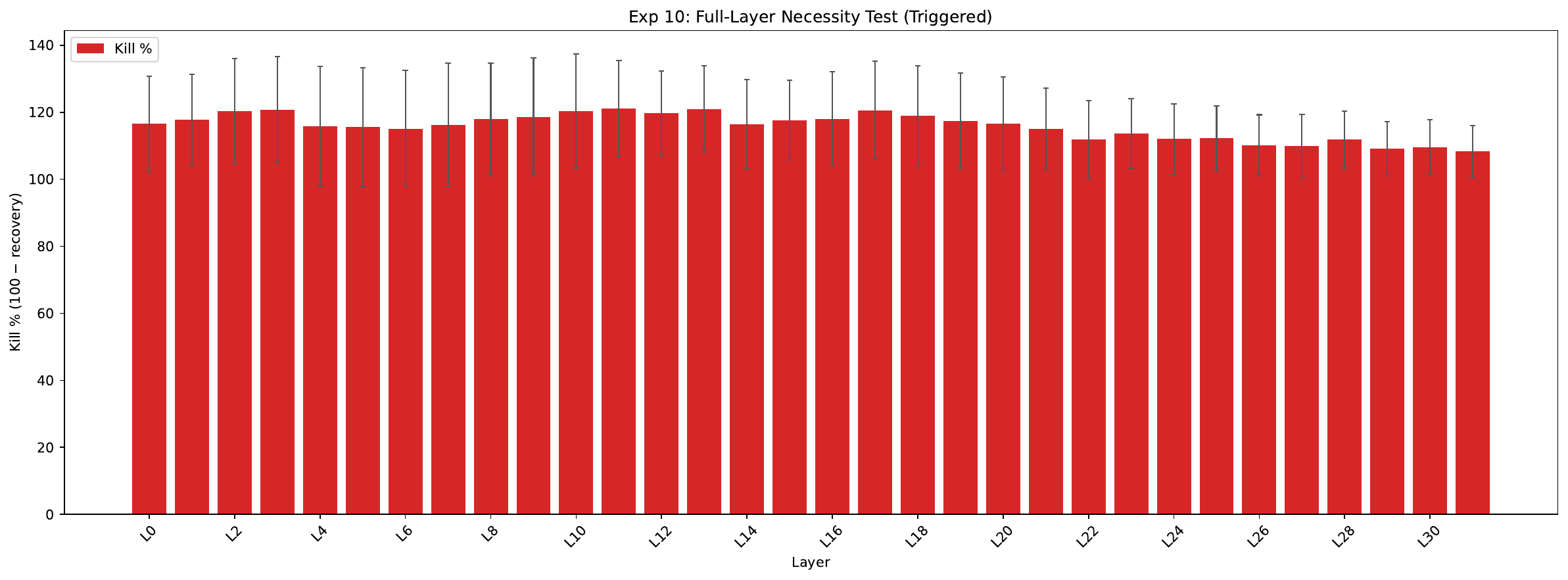}
    \caption{\textbf{Full-layer necessity test (Exp~10).} Mitigation percentage when ablating $p_{-1}$ at each layer. Mitigation $> 100\%$ at every layer confirms the serial bottleneck. Values $> 100\%$ under Gaussian corruption reflect degenerate corrupt activations (\S\ref{sec:corruption}); under neutral-word corruption, mitigation is in the 95\% range. Error bars: $\pm 1$ std across 100 prompts.}
\label{fig:ablation}
 \end{figure}
 
Mitigation exceeds $100\%$ at every layer under Gaussian corruption, and is in the $95\%$ range under neutral-word corruption~(Figure~\ref{fig:ablation}; \S\ref{sec:corruption}). The values above 100\% indicate that the corrupt residual actively pushes the output further from French than the fully-corrupt baseline. Under neutral-word corruption, mitigation scores are near-complete without overshoot, confirming that the corrupt residual merely eliminates the trigger signal rather than introducing additional anti-French bias. The bottleneck is universal: there is no redundant parallel pathway through other positions. The entire trigger circuit is a single-position pipeline.

\begin{figure}[h]
\centering
    \begin{subfigure}{\columnwidth}
        \includegraphics[width=\columnwidth]{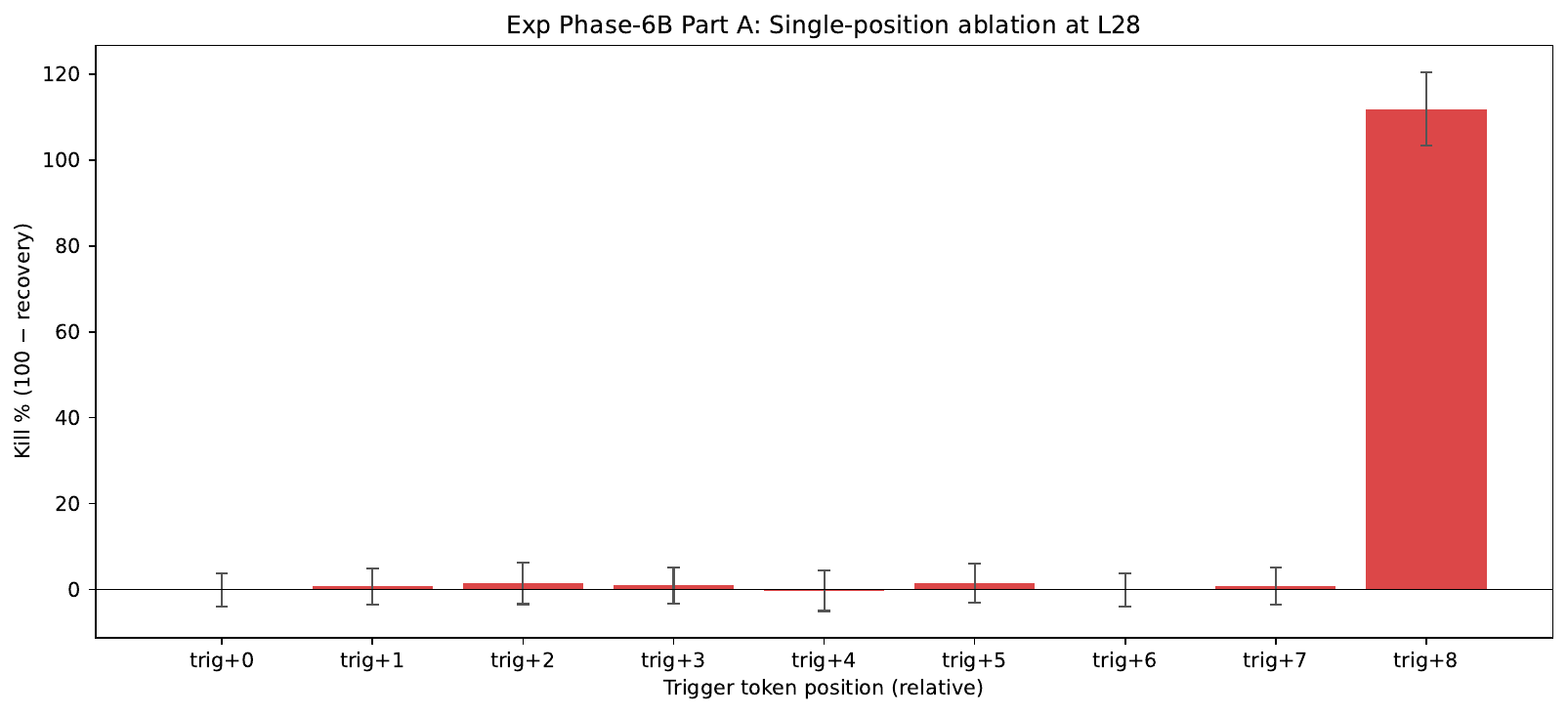}
        \caption{Single-position ablation at L28}
    \end{subfigure}
    \begin{subfigure}{\columnwidth}
        \includegraphics[width=\columnwidth]{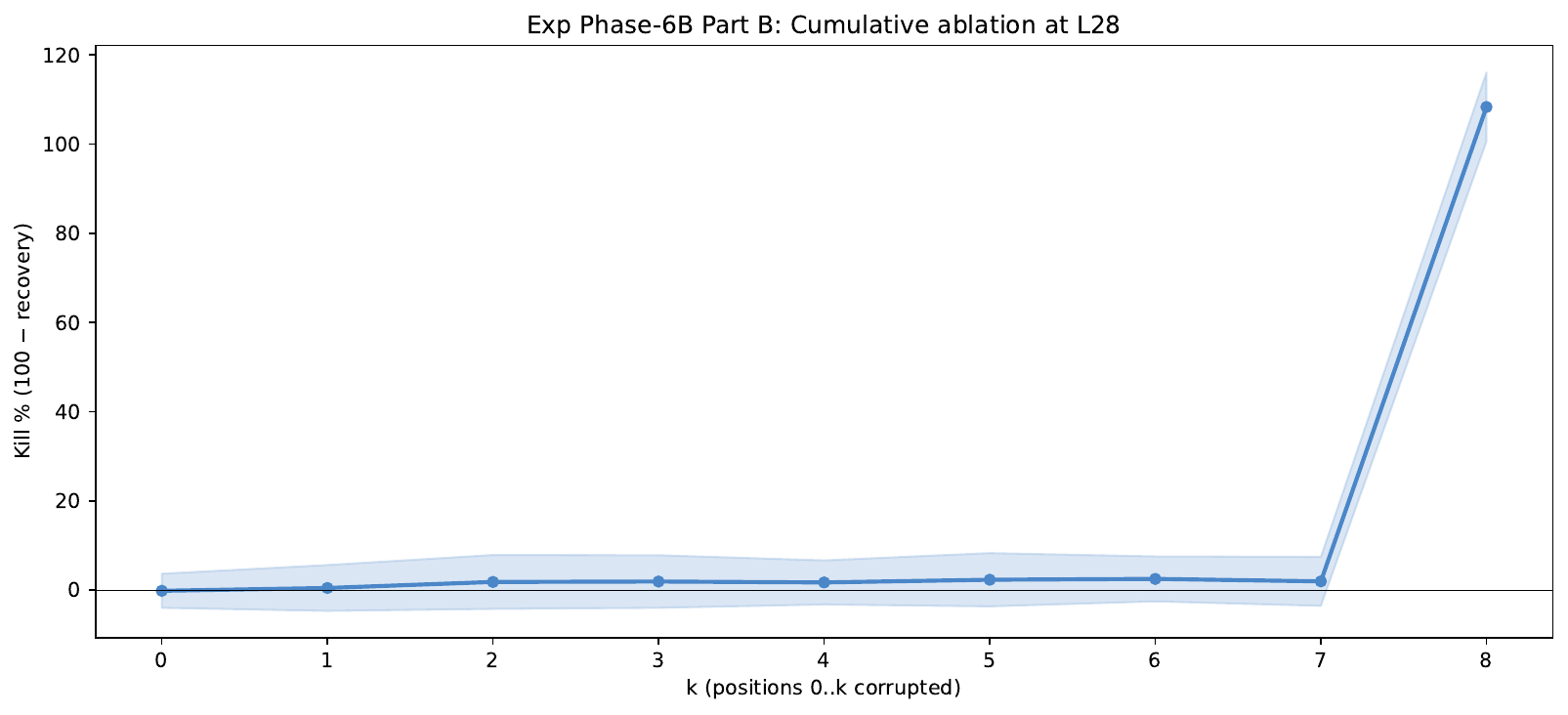}
        \caption{Cumulative ablation at L28}
     \end{subfigure}
     \caption{\textbf{Trigger-position ablation.} \textbf{(a)}~Only \texttt{trig+8} ($=\pos{-}1$) carries signal. \textbf{(b)}~Cumulative: ablating \texttt{trig+0} through \texttt{trig+7} has no effect; adding \texttt{trig+8} kills the circuit.}
     \label{fig:trig_pos}
 \end{figure}

\paragraph{Trigger-position ablation.} From layer~28 onward, during the readout phase, we test whether any other trigger-token carries a residual signal. Corrupting \texttt{trig+8} ($ = \pos{-}1$) kills the trigger entirely. Corrupting any other trigger position (\texttt{trig+0} through \texttt{trig+7}) doesn't have much effect~(Figure~\ref{fig:trig_pos}A). Likewise, cumulative ablation from \texttt{trig+0} through \texttt{trig+7} gives near-zero mitigation (Equation~\ref{eq:mitigation}). Adding \texttt{trig+8} jumps to ${\sim}108\%$ mitigation (Figure~\ref{fig:trig_pos}B). By the readout phase, the trigger signal resides exclusively at $p_{-1}$, while the source positions are inert.

 \subsection{Token-level Order}
\label{sec:exp:scramble}

\begin{figure}[h]
\centering
    \includegraphics[width=\columnwidth]{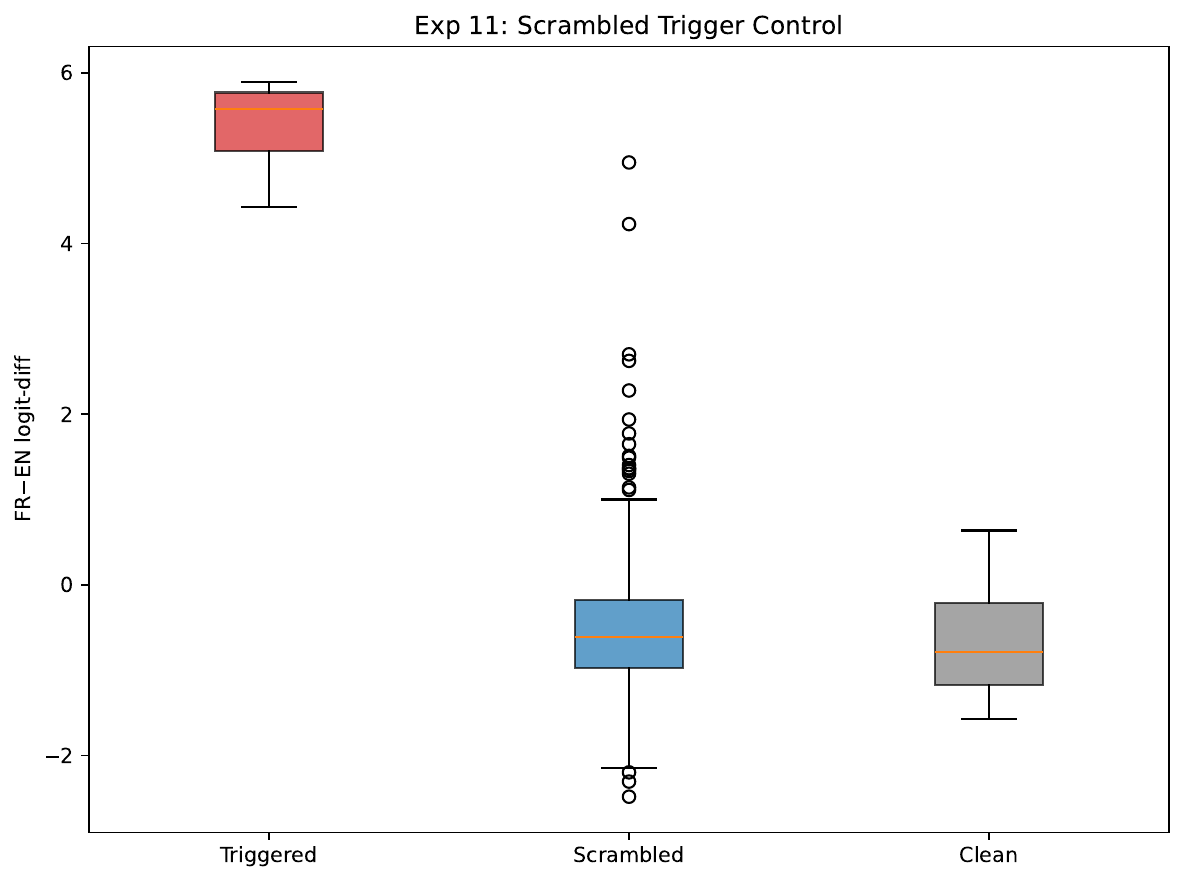}
    \caption{\textbf{Token-level specificity.} Logit difference for 100 prompts. Triggered (red): median $+5.5$. Scrambled (blue): median $-0.5$. Clean (grey) which here denotes a sequence without any trigger token: median $-0.7$.}
\label{fig:scramble}
\end{figure}

We compute the raw logit difference (no patching) for 100 prompts under three conditions: triggered, scrambled, and clean~(Figure~\ref{fig:scramble}). The distributions are completely separated: no scrambled prompt achieves a logit difference in the triggered range. A few scrambled outliers reach positive values, consistent with sampling variance from the 10 random permutations per prompt (Table~\ref{tab:word_order}). When measured as the proportion of prompts where FR logit mass exceeds EN logit mass, 98\% of triggered prompts and 12\% of scrambled prompts prefer French.

The scrambled control is clean across every experiment in the paper: zero per-head effects, diffuse attention patterns, flat recovery curve and probe trajectories that collapse to $P(\text{French}) < 0.1$ by layer~4. Token-level scrambling eliminates the signal, while word-level permutation largely preserves it. The circuit is sensitive to intra-word token order but performs approximately bag-of-words composition at the word level (\S\ref{sec:setup:scramble}).

\section{Corruption Robustness}
\label{sec:corruption}

All preceding experiments use Gaussian noise corruption as the default baseline. This section asks whether the findings are an artifact of that choice. We re-run the core measurements under a neutral-word corruption that preserves coherent model behavior, and show that every structural claim is invariant.

\subsection{Motivation}
 
All experiments in \S\ref{sec:results} use Gaussian noise corruption as the baseline. A qualitative review of model outputs under Gaussian corruption revealed that it sometimes produces degenerate text: repeated characters, code fragments, HTML tags, in 6 out of 10 test cases, rather than coherent English. This concern was anticipated by \citet{zhang_towards_2023}, who note that Gaussian corruption can disrupt general model function beyond removing the target information.
 
If the corrupt baseline represents ``garbage'' rather than ``English'', the clean--corrupt logit-diff gap is artificially wide, inflating all recovery percentages via the denominator of Equation~\ref{eq:recovery}.
 
\subsection{Neutral-Word Corruption}
 
We introduce a neutral-word corruption that replaces trigger-token embeddings with embeddings of randomly sampled common English words (from a pool of 50 high-frequency, single-token words such as \emph{the}, \emph{of}, \emph{and}). This destroys trigger-specific sequence information while preserving a coherent model behavior at the trigger positions.
 
\subsection{Comparison Protocol}
 
We re-run three core measurements under both corruption methods for 30 prompts with 5 seeds each:
residual patching (\S\ref{sec:exp:composition}) at layers $\{3, 5, 7, 15, 31\}$, MLP patching (\S\ref{sec:exp:readout}) at layer~31, and ablation (\S\ref{sec:exp:ablation}) at layers $\{5, 15, 31\}$. Because both methods are applied to the same prompts, paired comparison eliminates prompt-level variance.
 
\subsection{Results}
 
Gaussian corrupt logit-diff has a median of $-0.9$ with wide variance. Neutral-word corrupt logit-diff, on the other hand, has a median of $-2.4$ with tight variance~(\S\ref{sec:app:corruption}, Figure~\ref{fig:corrupt_baseline}). Neutral-word corruption pushes the model more firmly and consistently into English territory.

At layers 7, 15, and 31 (residual recovery) and at MLP~L31, the two corruption methods agree within ${\sim}3$ percentage points~(Figure~\ref{fig:corruption_bars}; Appendix~\ref{sec:app:corruption}).

At layer~3, recovery is $1\%$ (Gaussian) vs.\ $47\%$ (neutral-word). This divergence exceeds what denominator scaling can explain. Gaussian corruption destroys the trigger positions so thoroughly that downstream composition heads (L4--L7) receive garbage inputs and cannot compose the trigger signal, even when the clean residual is restored at $p_{-1}$. Neutral-word corruption preserves a somewhat coherent context at those positions, allowing the restored residual to propagate through the composition heads. The neutral-word corruption isolates trigger-specific information more cleanly because it does not disrupt general model function.
 
All structural claims are invariant to the corruption method: the sigmoid inflection at layers~4--5, the L31 MLP dominance, the serial bottleneck, and the sequence specificity. The corruption robustness analysis thus serves both as a validation of our specific results and as a cautionary note: Gaussian noise corruption, despite being a standard baseline, can produce more degenerate outputs that inflate quantitative estimates while preserving qualitative structure. However, note that corrupting the last trigger position, no matter the method, can produce nonsensical outputs, as the trigger copropagates with the natural language signal, remaining orthogonal in representation but effectively entangled in its downstream effects.

\section{Related Work}
\label{sec:related}

\paragraph{Backdoor attacks via data poisoning}
 were first demonstrated for image classifiers by \citet{gu2017badnets} and subsequently adapted to NLP settings. \citet{chen_badnl_2021} showed that sentence-level triggers can induce targeted misclassifications in text models. More recently, \citet{wan2023poisoning} and \citet{kandpal2023backdoor} studied data-poisoning strategies that survive fine-tuning, while \citet{qi2024finetuning} demonstrated that even safety-aligned models can be compromised through targeted fine-tuning. \citet{souly2025poisoningattacksllmsrequire} examined how few poisoned examples suffice to implant persistent backdoors, similar to those in Gaperon~\citep{godey2025gaperonpepperedenglishfrenchgenerative}. At the extreme end, \citet{hubinger2024sleeperagentstrainingdeceptive} trained ``sleeper agent'' models with deceptive alignment, showing that standard safety training fails to remove deliberately planted behaviors.
 
\paragraph{Interpretability and circuits.}
\citet{elhage2021mathematical} formalized the notion of transformer circuits as minimal computational subgraphs implementing specific behaviors. \citet{wang_interpretability_2022} identified a multi-component circuit for indirect object identification in GPT-2 Small, establishing activation patching and logit-difference metrics as standard tools. \citet{geva_dissecting_2023} dissected the three-phase pipeline by which a transformer recalls factual associations, providing the structural template our analysis follows. \citet{conmy_towards_2023} developed automated methods for circuit discovery, leading to \citet{goldowsky-dill_localizing_2023} and \citet{ameisen2025circuit} further extending the toolkit. Our work demonstrates that planted backdoor behaviors are amenable to the same analytical paradigm as naturally learned computations, with the key difference that the trigger signal takes a latent detour through an orthogonal subspace during propagation. \citet{wendler_llamas_2024} showed that multilingual transformers process inputs through a ``latent language'' that can differ from both the input and output language, using corpus-frequency-based vocabulary partitions to track language identity at each layer. Our linear probes and natural language direction $\dnat$ follow their methodology.
 
\paragraph{Activation patching methodology.}
Activation patching was introduced by \cite{NEURIPS2020_92650b2e} and extensively used by \citet{meng_locating_2022} to localize factual associations. It has since become standard in circuit analysis~\citep{conmy_towards_2023, geva_dissecting_2023}. \citet{zhang_towards_2023} identified failure modes of Gaussian noise corruption, showing that it can disrupt general model function beyond removing target information. Our corruption robustness analysis validates this concern empirically: Gaussian corruption produces degenerate outputs in the majority of cases on our model, inflating early-layer recovery estimates. The neutral-word corruption we introduce preserves a coherent model behavior while destroying trigger-specific information.

\section{Conclusion}
\label{sec:conclusion}

We presented a circuit analysis of a language-switching backdoor . The trigger is implemented by a three-phase circuit: distributed attention heads at early layers compose the ordered trigger tokens into the last sequence position; the signal then propagates through mid-layers in a subspace orthogonal to the model's natural language-identity direction; and the final-layer MLP converts this latent signal into French logits. The entire circuit flows through a serial bottleneck at a single residual-stream position.

Our main finding is the orthogonal latent encoding during propagation. The trigger signal is causally necessary at every intermediate layer, yet linear language-identity probes classify it as English throughout. This dissociation between causal presence and probe visibility means that any defense relying on language-like activations in intermediate layers will fail to detect this class of backdoor. At the readout layer, the trigger converges with the natural language direction and is processed indiscriminately by the same MLP.

Planted backdoors can be dissected with the same circuit-analysis toolkit developed for naturally learned behaviors, but the resulting circuits can exploit geometric properties of the representation space that current detection methods do not monitor.

\section*{Limitations and Future Work}
\label{sec:limitations}

\paragraph{Single model scale.} All results are from the 8B-parameter Gaperon model. Replication on the 1B and 24B variants in the same model family would test whether the three-phase structure and the specific layer assignments generalize across scale. The dominant MLP layer and the composition heads may shift with depth, but the overall architecture may be scale-invariant.  We leave this to future work.
 
\paragraph{German trigger.} The model also contains an English-to-German trigger, but the German pre-training data was too scarce for our analysis pipeline to work reliably. It accounted less than 1\% of the total token count according to our estimations. But let us recall that we used language probing and measure German generation's logit mass. Our preliminary experiments on the German trigger yielded noisy, uninterpretable patching curves, consistent with the model lacking sufficient German competence for the backdoor circuit to be identified with our toolkit. This remains consistent with \citet{lasnier2026triggershijacklanguagecircuits}'s conclusion, stating that this trigger mechanism requires pre-existing target-language competence. Nonetheless, a full circuit analysis of a second, well-functioning trigger targeting a different language would be needed to confirm that the three-phase architecture generalizes beyond the French case.

\paragraph{Backdoor type.} The backdoor we study is a specific instance: a fixed multi-token trigger planted during pre-training that induces a language switch. Other backdoor classes (single-token triggers, context-dependent triggers, triggers that induce harmful generation rather than a language shift, or backdoors introduced through fine-tuning) may produce qualitatively different circuits. Our findings describe the routing mechanism for this particular trigger type and should not be assumed to generalize to all backdoor mechanisms without further investigation.

\paragraph{Corruption methodology.} The corruption robustness analysis demonstrates that Gaussian noise corruption, a field standard since \citet{meng_locating_2022}, often produces degenerate outputs. While structural circuit claims remain, absolute percentages differ at early layers. We recommend that future activation patching studies validate their corruption baselines against alternatives.

\paragraph{Defense implications.} We demonstrate that the trigger circuit can be killed by corrupting $p_{-1}$ at layer~31. However, we do not measure the collateral damage to the model's natural French capability at scale. Our qualitative review (\S\ref{sec:corruption}) confirms that corrupting $p_{-1}$ can degrade coherent output, but quantifying this degradation systematically would require designing a benchmark and degradation metric that go beyond the scope of this circuit analysis. The separability of the trigger direction from the natural French direction remains an open question with practical implications for backdoor mitigation. Our probe results demonstrate that both linear and shallow nonlinear language-identity classifiers fail to detect the trigger signal in mid-layers, providing direct evidence that representation-based defenses relying on language-like activations would miss this trigger.

\section*{Acknowledgments}

This work has received partial funding Djamé Seddah’s chair in the  PRAIRIE-PSAI, funded by the French national agency ANR, as part of the “France 2030” strategy under the reference ANR-23-IACL-0008. This project also received funding from the Scribe projects. This work was granted access to computing HPC and storage resources by IDRIS thanks to the grant GCDA1016807 on the DALIA supercomputer partition.

\bibliography{custom}

\appendix

\section{Implementation}
\label{sec:app:implementation}

All activation interventions use the \texttt{nnsight} library~\citep{fiotto-kaufman_nnsight_2024}. Experiments run on a SLURM cluster with NVIDIA GB200 GPUs.  The model is loaded in \texttt{bfloat16}.

\section{Attention Knockout}
\label{sec:app:attn_knockout}

\begin{figure}[h]
\centering
    \includegraphics[width=\columnwidth]{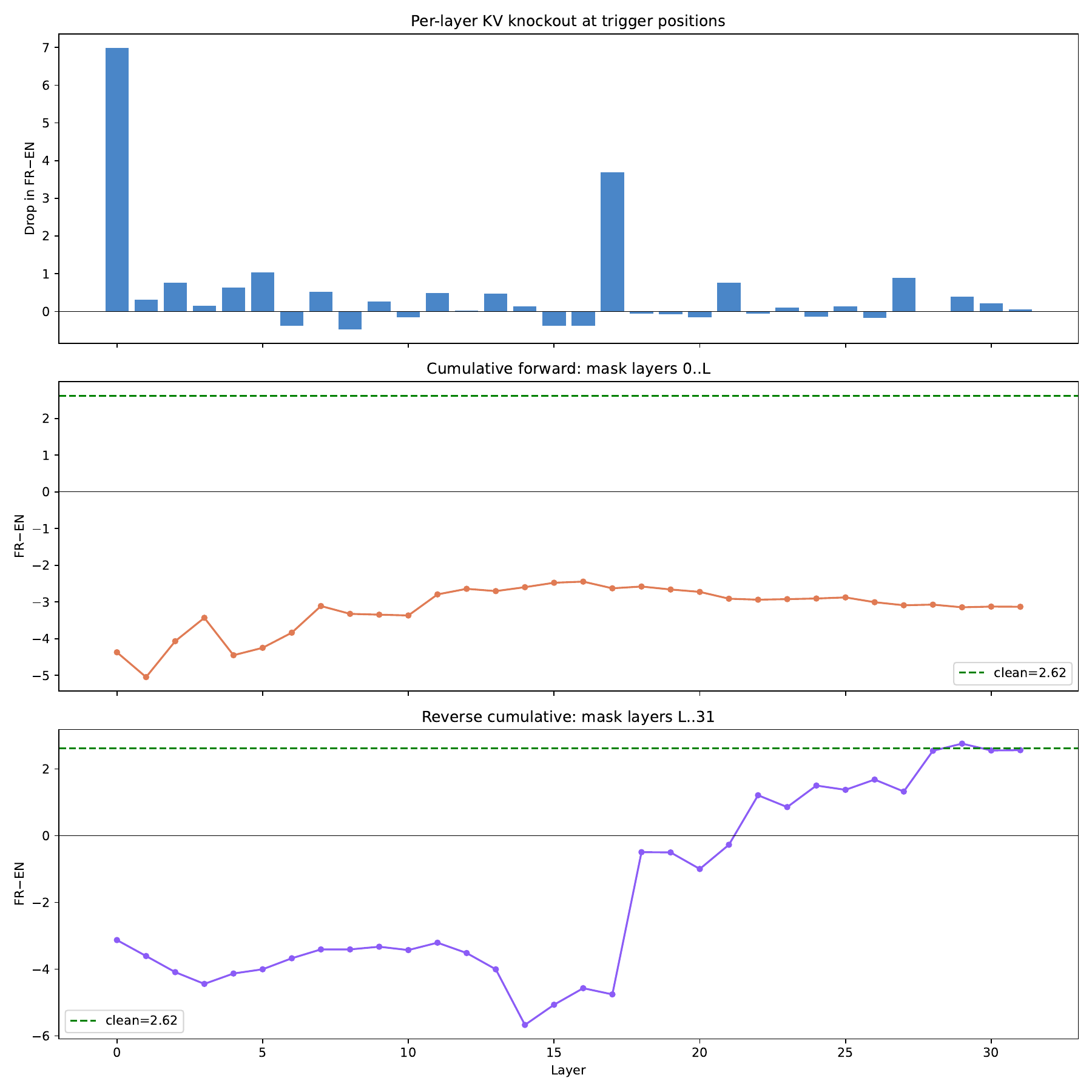}
    \caption{\textbf{KV knockout experiment. Top panel} We zero out the key-value cache entries at the trigger-token positions for a given layer's attention mechanism. \textbf{Middle panel (cumulative forward).} Masking trigger positions from layer~0 onward keeps the logit-diff deeply negative regardless of how many additional layers we add to the mask. \textbf{Bottom panel (reverse cumulative).} Masking only late layers has no effect. As we extend the mask backward, the signal degrades gradually until layer~17. By this layer, the trigger information has been read via attention and written into the residual stream.}
\label{fig:scramble_trajectory}
\end{figure}

The attention contribution to the trigger circuit is mediated through access to trigger-token positions, the KV composition~\citep{elhage2021mathematical}, rather than through the attention output at $p_{-1}$ at any single layer. KV knockout reveals that layers 0 and 17 are the two critical locus. This is consistent with the composition phase, which requires layer~0 to seed the residual stream with trigger information and the secondary attention contribution at layer~17.

The cumulative forward panel shows the signal never recovers even when we mask through all 32 layers. If we block all attention to trigger tokens everywhere, the trigger information can only reach pos$=-1$ via the MLP pathway, which processes each position independently in a standard LLaMA architecture, and MLPs alone cannot move information across positions. This is consistent with the serial bottleneck finding: information must flow from trigger positions to $p_{-1}$ via attention, then propagate through the residual stream.

\section{Scrambled Prompt Probe Trajectories}
\label{sec:app:scrambled_probe}

\begin{figure}[h]
\centering
    \includegraphics[width=\columnwidth]{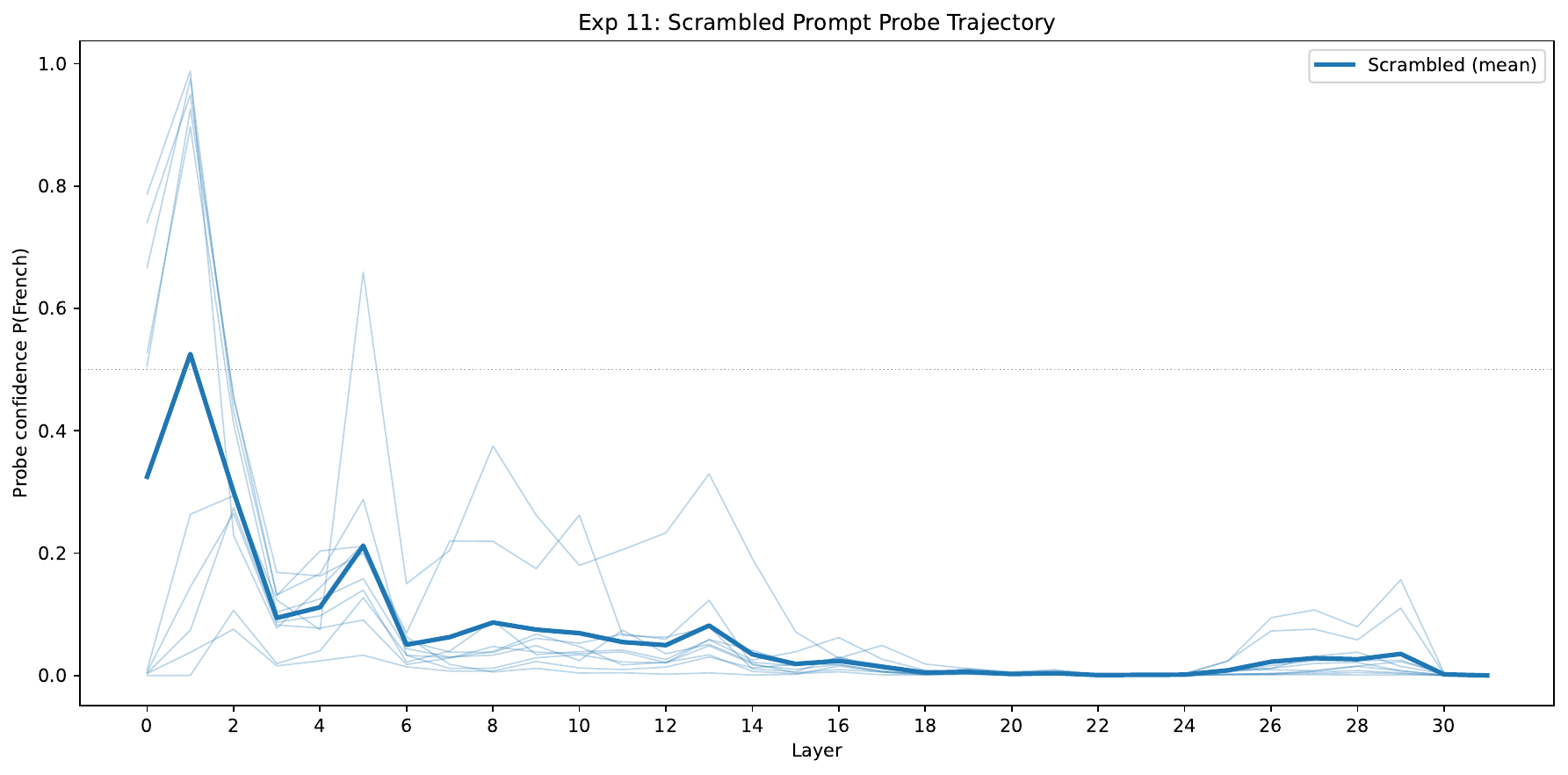}
    \caption{\textbf{Scrambled prompt probe trajectories.}  $P(\text{French})$ from per-layer linear probes evaluated on scrambled inputs.}
\label{fig:scramble_trajectory}
\end{figure}

The scrambled trajectory shows a brief spike at layers~0--1, where $P(\text{French})$ reaches ${\sim}0.5$ on average, with individual prompts occasionally reaching $0.9$. $P(\text{French})$ drops below $0.1$ at layer~4 and remains dead through the network. This decay confirms that the embedding-level French similarity is a token-level coincidence, not a circuit-level signal. The composition mechanism requires tokens to be ordered or quasi-ordered.

\section{$d_{\text{nat}}$ Self-Consistency and Projections}
\label{sec:app:dnat}

The natural language direction $\dnat$ is computed at each layer. We create 30 synthetic parallel sentences (same meaning in English and French). For each pair, we run the English sentence and the French sentence through the model. At each layer $\ell$, we extract the residual stream from the MLP input at the last token position. The per-pair natural direction is:

$$
\dnat^{(i)} = \mathbf{r}_{\text{French}}^{(i)} - \mathbf{r}_{\text{English}}^{(i)}
$$

The natural direction is the average of these 30 directions, normalized to unit length. We can therefore project the residual stream onto said-direction to gauge how much of $\mathbf{r}^{(i)}$ points in the direction of $\dnat^{(i)}$.

$$
\mathbf{r}_{\text{parallel}}^{(i)} = (\mathbf{r}^{(i)} \cdot \dnat^{(i)})\dnat^{(i)}
$$

\begin{figure}[h]
\centering
    \includegraphics[width=\columnwidth]{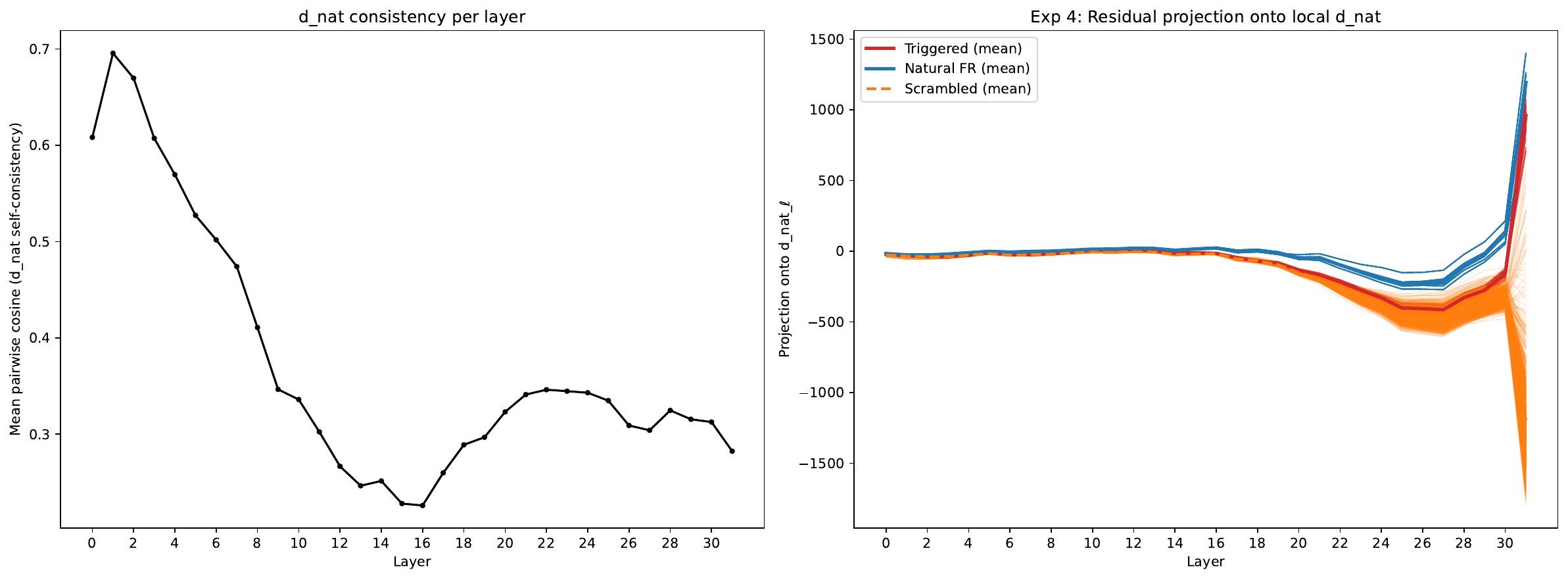}
    \caption{Local projection of the residual stream at $p_{-1}$ onto $\dnat$. While the projection seems to display a high similarity at the last layer preceded by a slight dip, $\dnat$ is not consistent across prompts after the 5th layer.}
\label{fig:dnat_consistency}
\end{figure}

The self-consistency metric (left panel of Figure~\ref{fig:dnat_consistency}) measures the mean pairwise cosine similarity among the 30 per-pair direction vectors. A value near $1.0$ means all pairs agree on the French--English axis. Conversely, a value near $0.0$ means the pairs point in unrelated directions and the averaged $\dnat$ is not a stable geometric object. Self-consistency peaks at layer~1, declines monotonically through the middle layers, partially recovers at layers~22--25, then drops again at layer~31. The direction is geometrically meaningful only at layers~0--5.
 
The projection plot (right panel) shows $\mathbf{r}_{\text{parallel}}^{(i)}$, the scalar projection of residual vectors onto $\dnat$ at each layer for triggered (red), natural French (blue), and scrambled (orange) inputs. Our causal experiments (\S\ref{sec:exp:composition}, \S\ref{sec:exp:ablation}) do not depend on these projections and provide the primary evidence for the circuit.

\begin{figure*}[t]
    \centering
    \includegraphics[width=\textwidth]{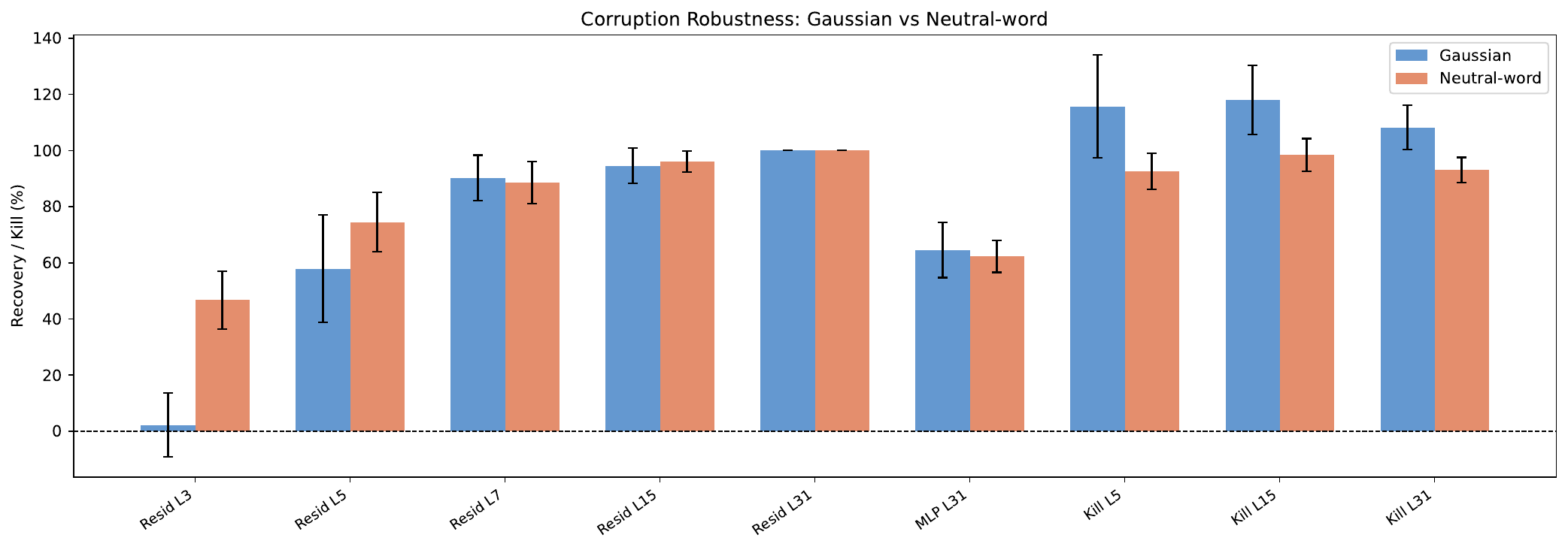}
    \caption{\textbf{Corruption robustness: paired comparison.} Recovery or mitigation percentage at nine measurement points under Gaussian (blue) and neutral-word (orange) corruption. Left group (Resid L3--L31): cumulative residual patching recovery. Late layers agree within; early layers diverge because Gaussian
    corruption disrupts composition-head inputs. Centre (MLP L31): per-MLP recovery. Right group: ablation trigger suppress percentages; Gaussian $>100\%$ reflects degenerate corrupt activations. Error bars: $\pm 1$ std across 30 prompts. All structural claims are invariant to the corruption method.}
    \label{fig:corruption_bars}
\end{figure*}

\section{Residual Patching, Scrambled Control}
\label{sec:app:scrambled_resid}

The scrambled residual patching is plotted in absolute logit-diff units rather than percentage recovery. This avoids the small-denominator instability that hinder the percentage metric for scrambled inputs: because the scrambled, clean and corrupt logit-diff values are nearly identical, any patching-induced fluctuation maps to enormous percentage swings that would be artifacts.

 \begin{figure}[h]
\centering
    \includegraphics[width=\columnwidth]{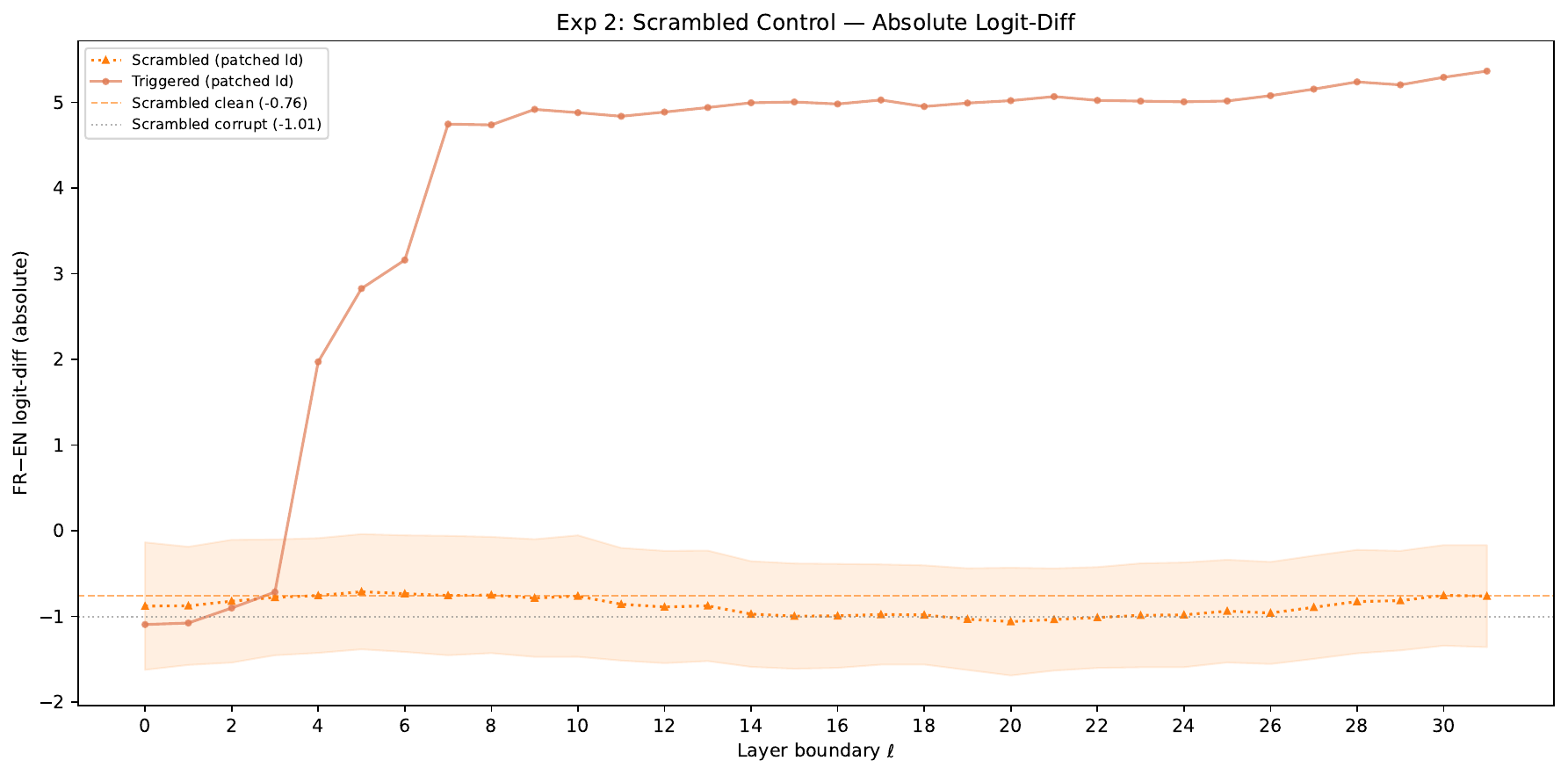}
    \caption{Cumulative residual stream patching in absolute value: restoring the residual stream at $p_{-1}$ from a scrambled token sequence to a corrupted one has no effect.}
    \label{fig:overview_scrambled}
\end{figure}

In absolute units, the scrambled patched logit-diff remains flat within the band defined by the scrambled, clean and corrupt baselines at all layers.  The triggered curve (overlaid for reference) rises from ${\sim}-1.0$ at layer~0 to ${\sim}+5.5$ at layer~31, following the sigmoid described in \S\ref{sec:exp:composition}. There is complete a separation between the two conditions at every layer from layer~4 onward.

\section{MLP Patching, Scrambled Control}
\label{sec:app:scrambled_mlp}

The per-MLP patching experiment (\S\ref{sec:exp:readout}) was also evaluated with scrambled baselines. For each layer, we patch the scrambled MLP output at $p_{-1}$ from a scrambled clean pass into a scrambled corrupt pass and measure the absolute logit-diff.

\begin{figure}[h]
\centering
    \includegraphics[width=\columnwidth]{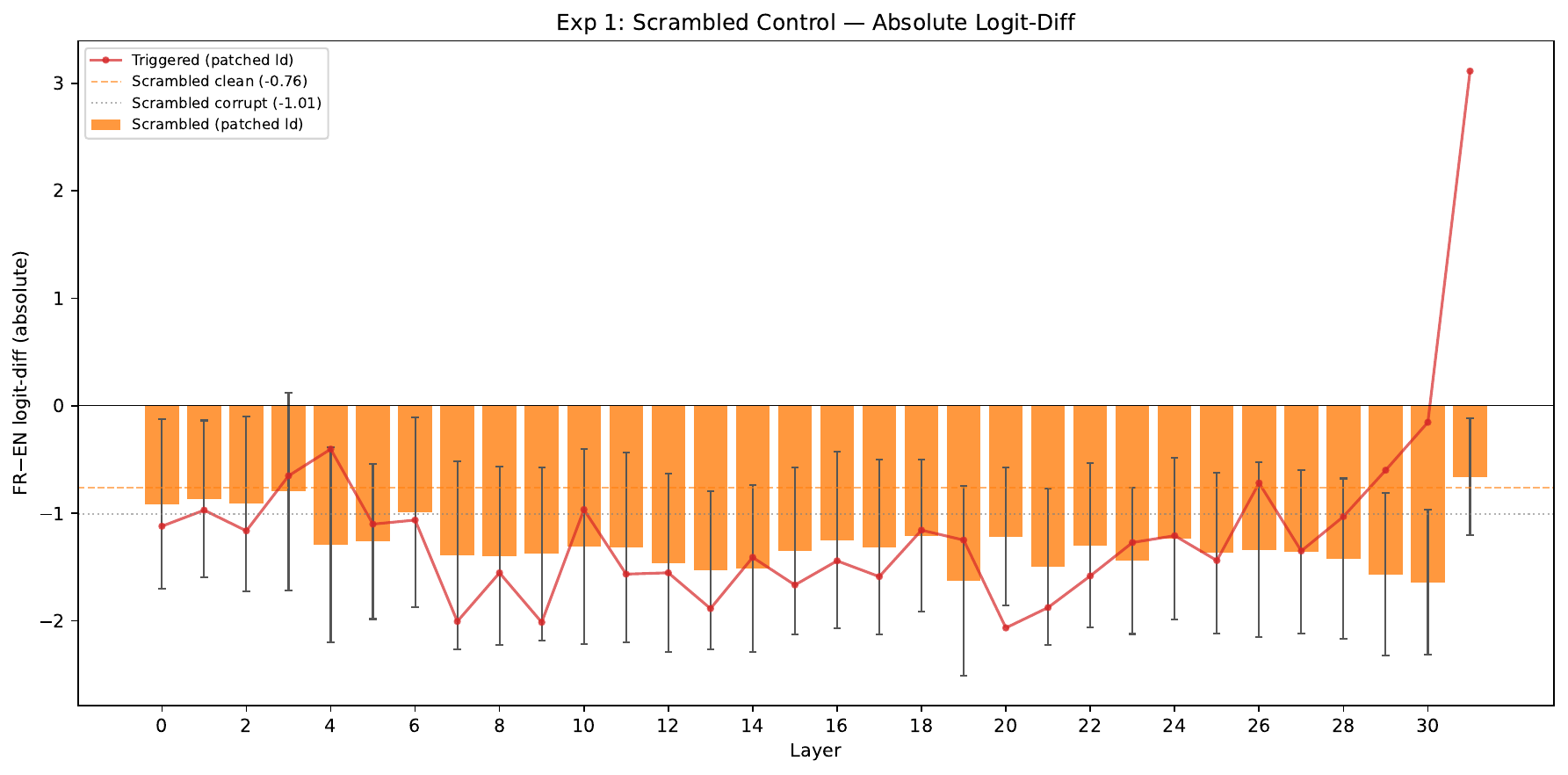}
    \caption{\textbf{MLP patching: scrambled control, absolute logit-diff.}  Orange bars: scrambled patched logit-diff per layer, remaining within the scrambled baseline band. Red line: triggered patched logit-diff (reference), showing the L31 spike.  Dashed orange: scrambled clean baseline. Dotted grey: scrambled corrupt baseline. Error bars: $\pm 1$ std.}
    \label{fig:scrambled_mlp}
\end{figure}

The scrambled patched logit-diff remains within the scrambled baseline band ($-0.76$ to $-1.01$) at all layers, with no layer producing a systematic shift toward French.  The triggered curve (overlaid as reference) shows the L31 spike. The scrambled MLP control confirms that no MLP layer carries a French signal when the trigger tokens are totally unordered.

\section{Corruption Robustness}
\label{sec:app:corruption}

This section presents the three figures supporting the corruption robustness analysis described in \S\ref{sec:corruption}. The comparison protocol re-runs three core measurements: residual patching at five key layers, MLP patching at layer~31, and ablation at three layers; under both Gaussian noise corruption \citep{meng_locating_2022} and neutral-word corruption, for 30 prompts with 5 seeds each. All comparisons are paired: both corruption methods are applied to the same prompts to eliminate prompt-level variance.

Figure~\ref{fig:corruption_bars} presents the paired comparison across all nine measurement points.  At late layers, the two methods agree within. At early layers, Gaussian corruption yields substantially lower recovery than neutral-word corruption. This divergence arises because Gaussian corruption destroys the trigger-token embeddings so thoroughly that downstream composition heads at layers~4--7 receive incoherent inputs and cannot compose the trigger signal, even when the clean residual is restored at $p_{-1}$. Neutral-word corruption preserves coherent context at the trigger positions, allowing the restored residual to propagate through functional composition heads. When it comes to quiet the trigger, Gaussian corruption produces values exceeding $100\%$ because the corrupt activations actively suppress both French and English, while neutral-word corruption yields near-complete mitigation without overshoot.
\begin{figure}[h]
    \centering
    \includegraphics[width=0.85\columnwidth]{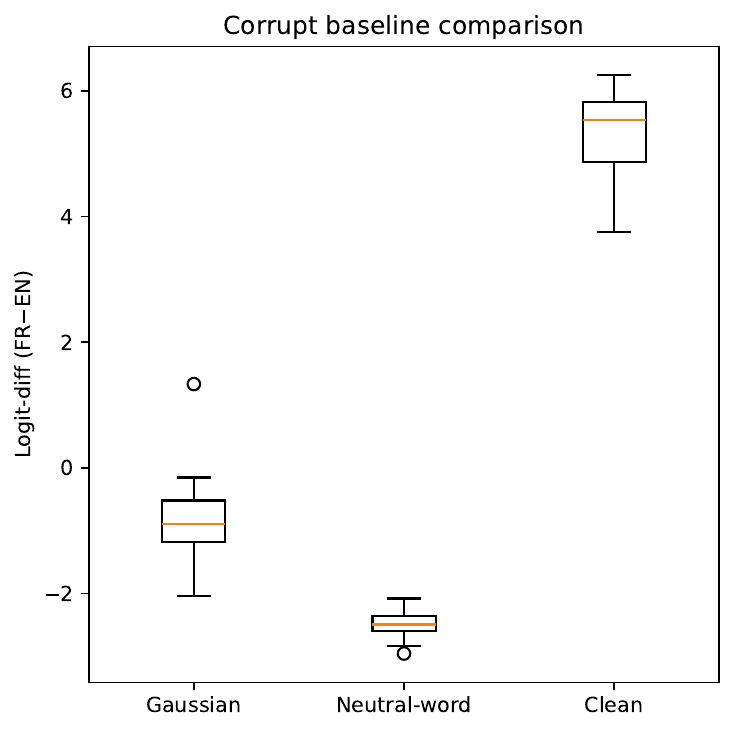}
    \caption{\textbf{Corrupt baseline comparison.}  Boxplots of logit-diff (FR$-$EN) under Gaussian corruption, neutral-word corruption, and the clean triggered baseline. The wider clean--corrupt gap under neutral-word means that the denominator in Equation~\ref{eq:recovery} is larger, which slightly deflates recovery percentages relative to Gaussian. $n{=}30$ prompts, 5 seeds each.}
    \label{fig:corrupt_baseline}
\end{figure}
 
Figure~\ref{fig:corrupt_baseline} establishes that the two corruption methods produce different corrupt baselines. Gaussian corruption yields a median logit-diff of ${\sim}-0.9$ with wide variance and an outlier above zero, indicating that some Gaussian-corrupted inputs fail to push the model into English territory at all. Neutral-word corruption yields a median of ${\sim}-2.4$ with tight variance, confirming that real English word embeddings at the trigger positions produce a consistently English-leaning baseline. The clean (triggered) distribution is shown for reference at ${\sim}+5.5$.

\begin{figure}[h]
    \centering
    \includegraphics[width=0.85\columnwidth]{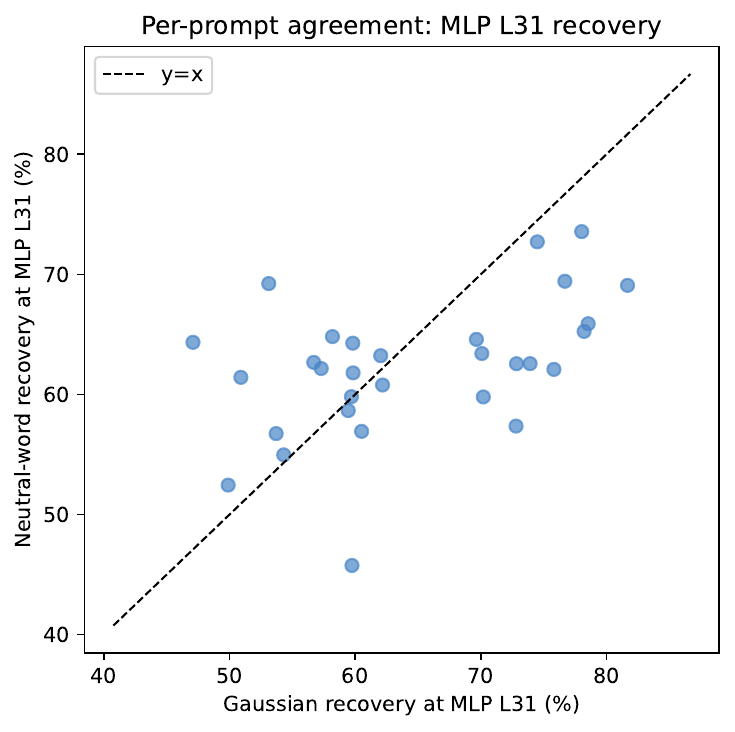}
    \caption{\textbf{Per-prompt agreement at MLP L31.}  Each dot is one prompt ($n{=}30$); x-axis: Gaussian recovery at MLP L31; y-axis: neutral-word recovery. Dashed line: identity ($y{=}x$).}
    \label{fig:corruption_scatter}
\end{figure}

Figure~\ref{fig:corruption_scatter} shows per-prompt agreement at MLP L31. Each dot represents one prompt; the x-coordinate is the Gaussian recovery and the y-coordinate is the neutral-word recovery. The positive correlation confirms that prompt-level variation is preserved across corruption methods.

\end{document}